\documentclass[lettersize,journal]{IEEEtran}
\usepackage{amsmath,amsfonts}
\usepackage{algorithmic}
\usepackage{algorithm}
\usepackage{array}
\usepackage{booktabs}
\usepackage{multirow}
\usepackage[caption=false,font=normalsize,labelfont=sf,textfont=sf]{subfig}
\usepackage{textcomp}
\usepackage{stfloats}
\usepackage{url}
\usepackage{verbatim}
\usepackage{graphicx}
\usepackage{cite}
\usepackage{xcolor}
\usepackage{amsmath}   
\usepackage{amssymb}    
\usepackage{bm}

\begin{document}

\title{MAPS: Multi-Anchor Projection Similarity for Joint Vision-Language Geo-Localization}

\author{Yutong Hu, Siyuan Tan, Shaocheng Yan, Pengcheng Shi, Qingwu Hu and Jiayuan Li
\thanks{Manuscript received xxx xx, xxxx; revised xxxx xx, xxxx. 
This work was supported in part by the Xiong'an New Area Science and Technology Innovation Special Program under Grant 2025XAGG0042, in part by the National Natural Science Foundation of China (NSFC) under Grant 42271444, and in part by the Postdoctoral Fellowship Program of CPSF under Grant Number GZB20250065.\textit{(Corresponding author: Jiayuan Li.)}

Yutong Hu, Siyuan Tan, Shaocheng Yan, Pengcheng Shi, Qingwu Hu and Jiayuan Li are with the
School of Remote Sensing and Information Engineering, Wuhan University,
Wuhan 430079, China, (e-mail: yutonghu@whu.edu.cn; ljy\textunderscore whu\textunderscore 2012@whu.edu.cn).
} }

\maketitle

\begin{abstract}
Humans localize places by integrating perceptual cues from vision with semantic reasoning from language, forming a scene understanding that is both intuitive and structured. Although existing geo-localization models have made substantial progress in cross-view and cross-modal settings, they are largely built upon point-to-point alignment, which is insufficient for joint vision-language queries. In such queries, visual and textual cues do not simply act as independent references, but jointly define a semantic subspace for locating the target. In this paper, we formulate vision-language geo-localization (VLGL) with joint image-text queries as a multi-anchor geometric alignment problem and propose a unified framework for this setting. To realize this formulation, we propose Multi-Anchor Projection Similarity (MAPS), a new metric which constructs an anchor plane from visual and textual query features in a high-dimensional space and measures similarity by the projection length of the target feature onto this plane. Unlike cosine similarity which evaluates isolated pairwise relations, MAPS captures the geometric consistency between the target feature and the joint query subspace, providing a more discriminative ranking criterion during retrieval. To make the learned representation consistent with this geometry, we further introduce a MAPS-based contrastive loss that drives target features toward the corresponding anchor plane. The proposed framework, similarity metric, and training objective jointly yield state-of-the-art performance in VLGL. Source code will be released at \url{https://github.com/YtH0823/MAPS}.
\end{abstract}


\section{Introduction}
\IEEEPARstart{H}{umans} infer the correspondence between limited observations and locations in the physical world, thereby determining ground position when GNSS is unavailable or unreliable \cite{lowry2015visual}. Geo-localization seeks to endow artificial agents with a comparable localization capability, which supports pedestrian navigation \cite{navigation}, robot positioning \cite{robot}, urban understanding \cite{VLAD} and emergency response \cite{disaster}. Beyond visual appearance matching, human spatial localization combines visual perception with textual semantic and spatial cues \cite{VLN,visualreasoning}. Visual cues reveal scene layout and environmental structure, while textual clues such as ``adjacent to the bridge" and ``facing the red building" provide relational constraints that help identify the target location.

Existing geo-localization studies have made substantial progress along two related directions. Cross-view geo-localization (CVGL) focuses on visual matching between ground-level imagery and aerial or satellite references \cite{transgeo,GeoDTR,Sample4geo,DAC,camp,dress}, while cross-modal geo-localization (CMGL) uses natural language descriptions to guide location retrieval \cite{text2pos,geotext1652,CORE,CVGText}. Contrastive learning provides an effective foundation for these settings by mapping heterogeneous observations into a shared embedding space and learning discriminative features from positive and negative sample relations \cite{contrastivelearning,infonce} \cite{cpc}. These advances establish a mature research basis for geo-localization and open the possibility of more flexible query forms.

However, when the query form extends to joint vision-language input, existing geo-localization methods still face a structural limitation. They largely inherit the pairwise alignment assumption of conventional contrastive learning, where each query cue is treated as an independent anchor and compared with the target through an isolated similarity score, typically cosine similarity \cite{CVMNET,Sample4geo}. This assumption is effective for single-query localization, but it overlooks the complementarity between visual and textual cues when they jointly describe the same place. Broader studies \cite{valor,VAST,GRAM,PMRL} in multimodal computer vision have explored alignment across more than two modalities, yet they mainly focus on general semantic fusion or global representation consistency, and are not tailored to the retrieval setting of vision-language geo-localization (VLGL). In VLGL, heterogeneous query cues must jointly constrain a geographically grounded target. Vision and language are therefore not independent references, but complementary anchors that define a shared semantic subspace in the embedding space, rather than a set of isolated cosine-based relations.

To address this structural mismatch, we introduce a multi-anchor geometric formulation for VLGL and instantiate it with UniMAG, a Unified Multi-Anchor Geo-localization framework. Unlike conventional pipelines that treat the query image and textual description as parallel inputs and optimize their pairwise alignments with candidate locations, UniMAG embeds the visual query, textual query, and candidate location features into a shared high-dimensional space. In this space, visual and textual cues act as coordinated anchors that jointly define the retrieval constraint, rather than independent signals that are compared with the target separately. This design provides a unified representation basis for VLGL, allowing the multi-anchor structure of joint queries to be explicitly modeled during both representation learning and retrieval ranking.

\begin{figure*}[!t]
\centering
\includegraphics[width=6.4 in]{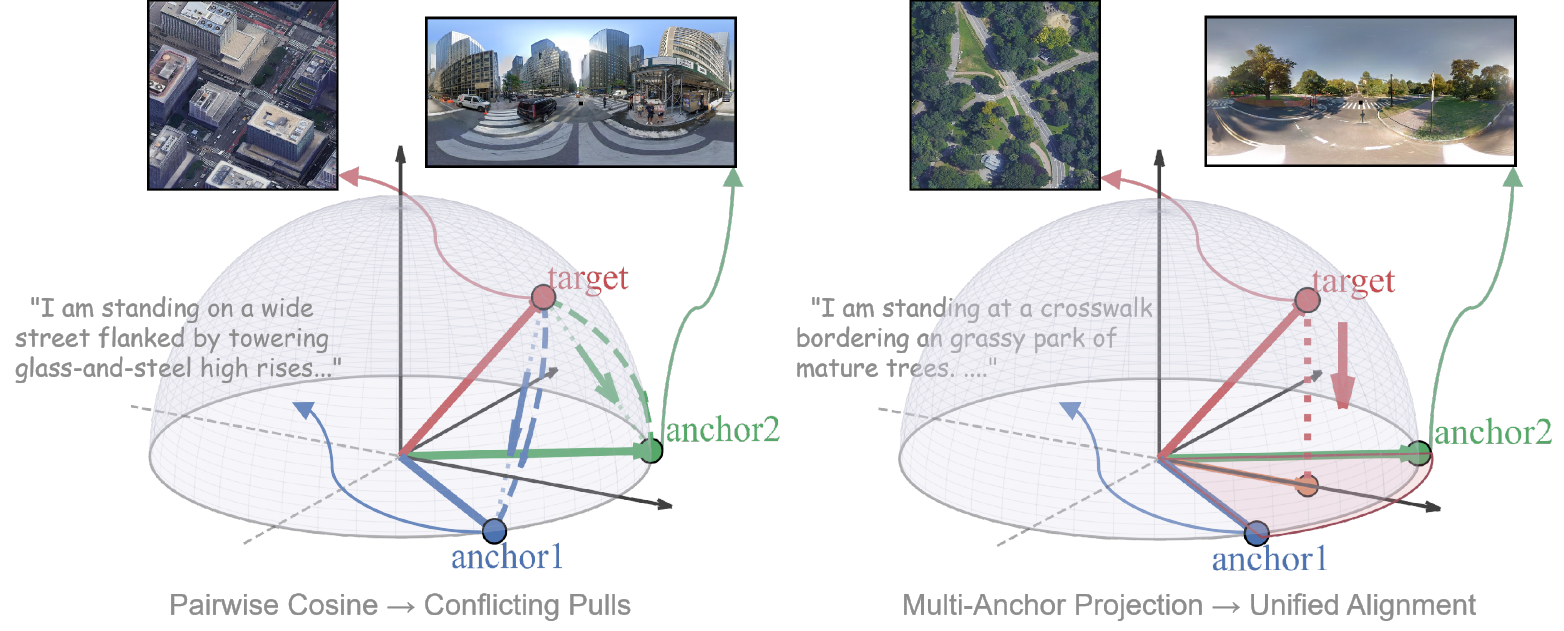}
\caption{Comparison between pairwise alignment and multi-anchor projection. Pairwise alignment optimizes point-to-point cosine distances, which pulls the target feature toward separate anchors and causes directional conflict. Multi-anchor projection guides the target feature toward the oriented plane spanned by the two anchors, yielding plane-consistent alignment with the joint query. }
\label{Motivation}
\end{figure*}

For alignment measurement in this multi-anchor context, we propose Multi-Anchor Projection Similarity (MAPS). Given the visual and textual query features, MAPS constructs an anchor plane in the shared embedding space and scores each candidate by the length of its projection onto this plane. For normalized features, this projection reflects the component of the candidate feature that is consistent with the two query anchors, so a larger value indicates stronger agreement with the joint vision-language constraint. To further reduce ambiguity caused by components outside the anchor plane, MAPS introduces an orientation-aware angular constraint derived from the relation between the candidate feature and the directed anchor plane. This constraint penalizes out-of-plane deviations and favors candidates that better conform to the vision-language-consistent subspace. As illustrated in Fig.~\ref{Motivation}, MAPS measures the geometric consistency between a candidate feature and the query plane spanned by multiple anchors. It shifts retrieval ranking from standard cosine-based pairwise matching to multi-anchor projection, which avoids imposing conflicting alignment directions on the target feature and provides a more structured similarity criterion for VLGL.

Beyond replacing cosine similarity with a geometry-aware retrieval metric, MAPS further leads to a contrastive loss that makes representation learning consistent with the proposed multi-anchor formulation. Rather than optimizing positive and negative samples only through isolated pairwise similarity, this MAPS-based loss encourages positive candidate features to produce stronger projection responses on the corresponding anchor plane, while suppressing negatives that are inconsistent with the joint query geometry. In this way, the network learns features that are not only discriminative in the shared embedding space, but also better aligned with the vision-language subspace used for retrieval.

We conduct extensive experiments on two geo-localization datasets that provide both ground-level imagery and textual descriptions, including CORE \cite{CORE} and CVG-Text \cite{CVGText}. Across these two datasets, our proposed UniMAG consistently improves localization accuracy across different regions and reference modalities, with larger gains observed when MAPS is used for both retrieval ranking and representation learning. These results empirically support our formulation of VLGL as a geometric alignment problem over a joint query subspace, rather than a set of isolated pairwise comparisons. Our contributions can be summarized as follows:

\begin{itemize}
\item We formulate VLGL as a multi-anchor geometric alignment problem and propose UniMAG, a unified framework that embeds visual queries, textual queries, and candidate locations into a shared high-dimensional space, where heterogeneous cues jointly define the retrieval constraint.

\item We propose Multi-Anchor Projection Similarity (MAPS), a target-to-subspace similarity metric designed for joint vision-language queries. MAPS builds an anchor plane from visual and textual query features and ranks each candidate by its projection consistency with this plane, replacing isolated cosine comparisons with geometry-aware multi-anchor matching.

\item We further introduce a MAPS-based contrastive loss that optimizes feature learning under the same multi-anchor geometry used for retrieval. By encouraging positive candidates to align with the query-defined anchor plane and suppressing negatives outside the joint constraint, this objective improves the consistency between training and inference.
\end{itemize}

\section{Related Work}

\subsection{Geo-localization Task}

Geo-localization retrieves target locations from a geo-referenced database according to query information. Existing studies mainly follow two paradigms. CVGL matches ground-view images with overhead imagery and formulates geo-localization as a cross-view visual retrieval problem \cite{CVUSA,CVACT,vigor,CVCITIES,CVGlobal,DAC,dress}. Methods such as Sample4Geo \cite{Sample4geo} show that contrastive representation learning effectively reduces the viewpoint gap between ground-view and satellite images. CMGL \cite{geotext1652,geobridge,CVGText} further extends the query form from images to language, where methods such as PLANET \cite{CORE} align scene descriptions with satellite imagery through vision-language representation learning. Despite their progress, both CVGL and CMGL largely follow a single-anchor retrieval formulation, where the query is usually either a ground-view image or a textual description. This formulation overlooks practical localization scenarios in which visual observations and language descriptions jointly constrain the same target location \cite{anderson2018vision,visualreasoning,touchdown}. Our proposed VLGL task addresses this limitation by modeling geo-localization as joint vision-language retrieval, where image queries, text queries, and geo-referenced candidates are represented in a shared embedding space for unified location matching.

\begin{figure*}[!t]
\centering
\includegraphics[width=\textwidth]{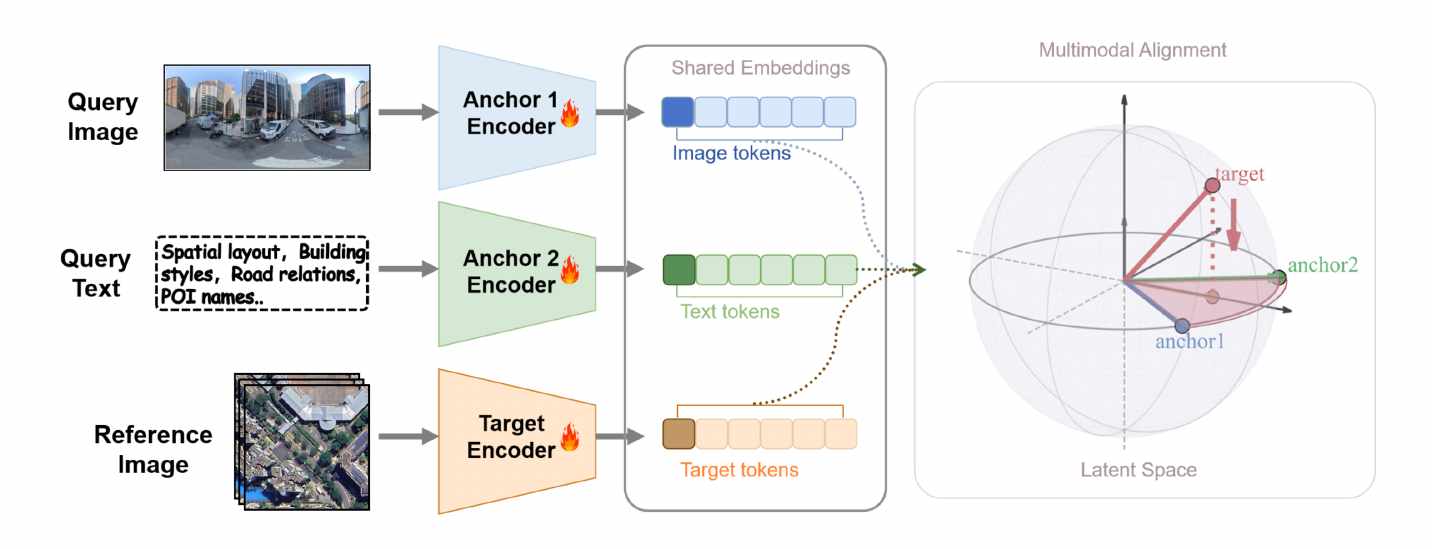}
\caption{Pipeline of the UniMAG framework. UniMAG encodes the ground-view image query, textual query, and geo-referenced candidates into a shared embedding space, where multimodal representations are aligned. By jointly modeling the relations among visual queries, language queries, and reference candidates, UniMAG formulates VLGL as geo-referenced retrieval with joint image-text queries. This framework provides a unified representation basis for subsequent multi-anchor similarity modeling.}
\label{fig:maps_pipeline}
\end{figure*}

\subsection{Multimodal Alignment}

Multimodal alignment aims to represent heterogeneous signals in a shared space where semantically related instances can be matched across modalities \cite{multimodal1,multimodal2,multimodal3,IR,IR2}. Early progress is mainly driven by bimodal contrastive learning, where vision-language models such as CLIP \cite{EVAClip,CLIP,blip,remoteclip,longclip,tagclip,vsepp,scan,Vilbert,Visualbert,Uniter,albef} align image and text representations through pairwise similarity. This paradigm is later extended to other modality pairs, including audio-text \cite{audio2text}, video-text \cite{video2text}, and point-text alignment \cite{point2text}, and provides the foundation for many multimodal retrieval systems. Recent multimodal foundation models further incorporate more modalities into a unified framework. For example, ImageBind \cite{imagebind} aligns multiple modalities by using one modality as the anchor, while VAST \cite{VAST} jointly models vision, audio, subtitles, and text with additional multimodal objectives. Despite their effectiveness, these methods still largely depend on pairwise alignment or a predefined anchor modality, which makes the relations among non-anchor modalities implicit and limits the modeling of joint multimodal structure. Recent studies attempt to move beyond pairwise alignment by directly modeling multiple modality representations as a whole. GRAM \cite{GRAM} introduces a Gramian volume measure to align multiple modalities simultaneously, while PMRL \cite{PMRL} further connects full multimodal alignment with the rank of the Gram matrix and optimizes the dominant singular direction. These works show that multimodal relations can be better characterized by higher-order geometric structures rather than isolated pairwise similarities. However, VLGL differs from general multimodal alignment. In our setting, visual and textual cues are not independent modalities to be collapsed into a single representation, but complementary query anchors that jointly constrain a candidate location. Therefore, our MAPS formulation focuses on target-to-subspace consistency, measuring how well a geo-referenced candidate aligns with the visual-textual query subspace instead of enforcing global modality collapse.

\section{Methodology}

\subsection{Problem Formulation}

Formally, VLGL is cast as a joint-query location retrieval problem. Given a query set $\mathcal{Q}=\{(G_i,T_i)\}_{i=1}^{M}$, where $G_i$ denotes a ground-level query image and $T_i$ denotes its associated textual description, and a gallery of $N$ geo-tagged reference candidates $\mathcal{R}=\{R_1,R_2,\dots,R_N\}$, the objective is to retrieve the reference candidate corresponding to the physical location of a specific joint query $Q=(G,T)$. Depending on the retrieval setting, these reference candidates may comprise satellite imagery, map tiles, or other geo-referenced representations.

Let $\Phi_{vis}(\cdot)$, $\Phi_{txt}(\cdot)$, and $\Phi_{ref}(\cdot)$ denote the visual, textual, and reference encoders, respectively. These functions map cross-modal inputs into a shared $d$-dimensional latent space:
\begin{equation}
\mathbf{v}=\Phi_{vis}(G), \quad \mathbf{t}=\Phi_{txt}(T), \quad \mathbf{r}_k=\Phi_{ref}(R_k).
\end{equation}

The retrieval objective is thus defined as:
\begin{equation}
\hat{R}=\mathop{\arg\max}_{R_k \in \mathcal{R}} \operatorname{Sim}_{vl}(\mathbf{r}_k; \mathbf{v}, \mathbf{t}),
\end{equation}
where $\operatorname{Sim}_{vl}(\cdot)$ measures the semantic compatibility between a candidate location feature and the joint vision-language query.

Conventional formulations typically decouple this compatibility into independent pairwise similarities \cite{CLIP,CVMNET,Sample4geo,CVGText,CORE}. While straightforward and effective for single-source retrieval, this paradigm treats visual and textual queries as isolated anchors, failing to explicitly model the intrinsic semantic structure jointly defined by them. In the context of VLGL, an optimal similarity function must evaluate whether a candidate feature is geometrically consistent with the joint query subspace, rather than merely assessing its marginal proximity to individual query features. This limitation directly motivates our multi-anchor formulation and the subsequent architectural design of UniMAG and MAPS.

\subsection{UniMAG Framework}
In this paper, we propose UniMAG, a unified multi-anchor geo-localization framework that provides a consistent representation basis for joint vision-language retrieval. UniMAG is designed as a representation-oriented framework that projects heterogeneous observations into a shared high-dimensional embedding space, providing a unified basis for subsequent multi-anchor geometric modeling. As illustrated in Fig.~\ref{fig:maps_pipeline}, given a ground-level query image, its associated textual description, and a geo-tagged reference candidate, UniMAG uses modality-specific encoders to generate the corresponding feature vectors \cite{CLIP,blip,imagebind}, denoted as $\mathbf{v}$, $\mathbf{t}$, and $\mathbf{r}$, respectively. To ensure a consistent feature distribution on the common hypersphere, all feature vectors are normalized by $L_2$ normalization.

Within this shared space, the visual and textual features are no longer treated as isolated query signals for independent alignment. They instead serve as coordinated anchors that provide the geometric basis for constructing the joint query subspace. This design explicitly preserves the semantic complementarity between vision and language, and provides the necessary embedding environment for the multi-anchor similarity calculation introduced in the following section.

\begin{figure*}[!t]
\centering
\includegraphics[width=\textwidth]{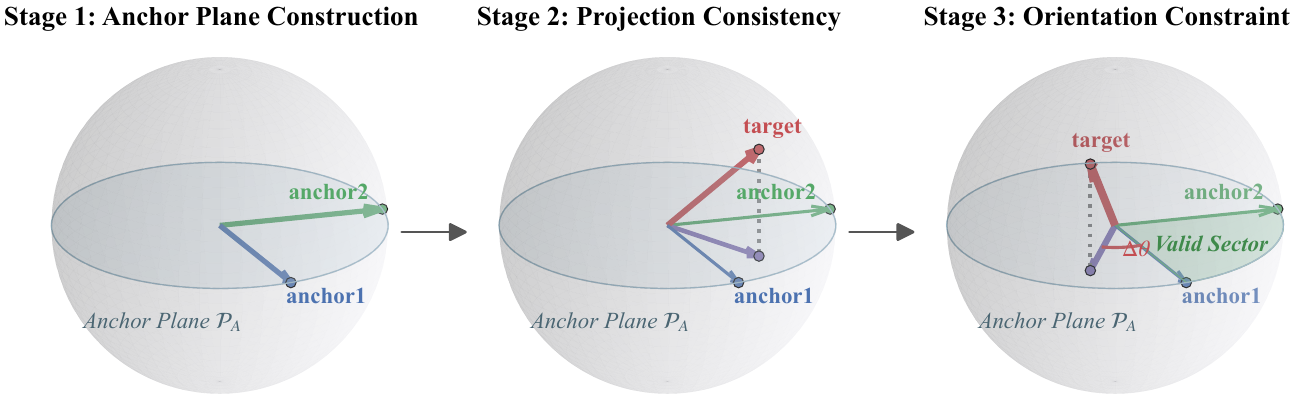}
\caption{Computation process of MAPS. MAPS first constructs a query anchor plane from the visual and textual anchors, and then projects the candidate reference feature onto this plane to measure target-to-subspace consistency. An orientation-aware constraint is further introduced to penalize projected directions outside the valid visual-textual query sector. The final score jointly captures projection consistency and directional validity.}
\label{fig:maps_computation}
\end{figure*}

\subsection{Multi-Anchor Projection Similarity}

The computation process of MAPS is summarized in Fig.~\ref{fig:maps_computation}.

\subsubsection{Gramian Geometry Theory}

Building on the shared embedding space of UniMAG, MAPS studies the geometric relation among the visual query, the textual query, and each candidate reference. The visual query feature, textual query feature, and candidate feature are denoted as $\mathbf{v}\in\mathbb{R}^{d}$, $\mathbf{t}\in\mathbb{R}^{d}$, and $\mathbf{r}\in\mathbb{R}^{d}$, respectively. Within the normalized feature space \cite{CLIP,Sample4geo}, these embeddings lie on the unit hypersphere, satisfying $\|\mathbf{v}\|_2=\|\mathbf{t}\|_2=\|\mathbf{r}\|_2=1$. Instead of reducing the joint query to two independent reference signals, MAPS regards $\mathbf{v}$ and $\mathbf{t}$ as coordinated anchors that define the geometric support of the visual-textual query. The visual and textual query features are collected into an anchor matrix $\mathbf{A}=[\mathbf{v},\mathbf{t}]\in\mathbb{R}^{d\times 2}$, whose column space defines the query-induced anchor plane $\mathcal{P}_{\mathbf{A}}=\operatorname{span}(\mathbf{A})$ when the two anchors are not collinear. This plane provides the geometric basis for evaluating whether a candidate reference is consistent with the joint visual-textual constraint. The intrinsic geometry of this plane is described by the Gram matrix of the two anchors:
\begin{equation}
\mathbf{G}_{\mathbf{A}}
=
\mathbf{A}^{\top}\mathbf{A}
=
\begin{bmatrix}
\mathbf{v}^{\top}\mathbf{v} & \mathbf{v}^{\top}\mathbf{t}\\
\mathbf{t}^{\top}\mathbf{v} & \mathbf{t}^{\top}\mathbf{t}
\end{bmatrix}
=
\begin{bmatrix}
1 & S_{vt}\\
S_{vt} & 1
\end{bmatrix},
\end{equation}
where $S_{vt}=\mathbf{v}^{\top}\mathbf{t}$ denotes the cosine similarity between the visual and textual anchors. According to Gramian geometry, which has been recently explored for multimodal learning \cite{GRAM,PMRL}, the determinant of this matrix equals the squared area of the parallelogram spanned by the two anchors:
\begin{equation}
\det(\mathbf{G}_{\mathbf{A}})
=
1-S_{vt}^{2}.
\end{equation}
Here, $\det(\cdot)$ denotes the determinant operator. This determinant is not used as a standalone evaluation score. Instead, it characterizes the geometric extent of the anchor plane in the feature space, thereby providing the basis for measuring the subsequent alignment between a candidate feature and the joint query subspace. The following derivation assumes $\det(\mathbf{G}_{\mathbf{A}})>0$, which means that the two anchors define a non-degenerate plane; otherwise, the anchor plane collapses to a lower-dimensional subspace. To characterize how a candidate feature relates to the anchor plane, the candidate feature $\mathbf{r}$ is further incorporated into the Gramian geometry as $\mathbf{B}=[\mathbf{v},\mathbf{t},\mathbf{r}]\in\mathbb{R}^{d\times 3}$. Applying the same Gramian construction to this augmented matrix yields
\begin{equation}
\mathbf{G}_{\mathbf{A},\mathbf{r}}
=
\mathbf{B}^{\top}\mathbf{B}
=
\begin{bmatrix}
1 & S_{vt} & S_{vr}\\
S_{vt} & 1 & S_{tr}\\
S_{vr} & S_{tr} & 1
\end{bmatrix},
\end{equation}
where $S_{vr}=\mathbf{v}^{\top}\mathbf{r}$ and $S_{tr}=\mathbf{t}^{\top}\mathbf{r}$. According to the Gram determinant interpretation, $\det(\mathbf{G}_{\mathbf{A},\mathbf{r}})$ gives the squared volume of the parallelotope spanned by $\mathbf{v}$, $\mathbf{t}$, and $\mathbf{r}$. If $\mathbf{r}$ lies in the anchor plane defined by $\mathbf{v}$ and $\mathbf{t}$, the three vectors become linearly dependent, leading to $\det(\mathbf{G}_{\mathbf{A},\mathbf{r}})=0$. If $\mathbf{r}$ deviates from the anchor plane, the volume increases with the orthogonal component that cannot be explained by the joint query subspace. The normalized relation between this augmented Gramian volume and the squared area of the anchor plane therefore provides the geometric basis for deriving MAPS as a projection-based similarity measure.

\subsubsection{MAPS as Projection Consistency}

In a normalized feature space, the conventional cosine similarity has a simple geometric interpretation in single-anchor retrieval \cite{CLIP,CVMNET,Sample4geo}. Given a unit query anchor $\mathbf{a}$ and a unit candidate feature $\mathbf{r}$, cosine similarity is the signed scalar projection of $\mathbf{r}$ onto the anchor direction:
\begin{equation}
s_{\cos}(\mathbf{r},\mathbf{a})
=
\mathbf{a}^{\top}\mathbf{r}.
\end{equation}
Thus, cosine similarity can be regarded as an orientation-sensitive projection response along a single query direction. This single-anchor interpretation motivates our extension from marginal cosine projections to projection consistency with the visual-textual anchor plane.

Although existing multimodal alignment methods learn cross-modal consistency through global or high-order constraints, joint queries in VLGL follow a more explicit asymmetric retrieval structure. In this setting, the visual feature $\mathbf{v}$ and textual feature $\mathbf{t}$ serve as query anchors, and the candidate feature $\mathbf{r}$ is evaluated by its consistency with the query-induced geometry. MAPS therefore introduces a task-specific geometric criterion by constructing an anchor plane from the visual and textual queries and measuring the candidate through its projection consistency with this plane. This formulation explicitly separates the geometry of the query anchors from the candidate's orthogonal deviation, yielding a similarity measure better suited to VLGL.

Formally, let the anchor plane spanned by the visual and textual query features be denoted as $\mathcal{P}$. The orthogonal distance from the candidate feature $\mathbf{r}$ to this plane can be explicitly derived using the Gram determinant. The Gram matrix of a set of vectors computes their pairwise inner products, and its determinant quantifies the squared volume of the parallelepiped they form. Thus, the squared orthogonal distance from $\mathbf{r}$ to the subspace $\mathcal{P}$ is given by the ratio of the squared volume of the parallelepiped formed by all three vectors to the squared area of the parallelogram formed by the query anchors:
\begin{equation}
d^{2}(\mathbf{r}, \mathcal{P}) = \frac{\det(\mathbf{G}_{\mathbf{A},\mathbf{r}})}{\det(\mathbf{G}_{\mathbf{A}})}
\end{equation}
Since the candidate feature $\mathbf{r}$ resides on the unit hypersphere, its squared norm is strictly equal to one. By applying the Pythagorean theorem, the magnitude of the projection of $\mathbf{r}$ onto the anchor plane $\mathcal{P}$, which serves as the base formulation for our Multi-Anchor Projection Similarity, is computed as:
\begin{equation}
s_{\text{proj}}(\mathbf{r}, \mathbf{v}, \mathbf{t}) = \sqrt{1 - \frac{\det(\mathbf{G}_{\mathbf{A},\mathbf{r}})}{\det(\mathbf{G}_{\mathbf{A}})}}
\end{equation}
This projection magnitude naturally extends the concept of cosine similarity to a multi-anchor setting. A larger projection length implies that the candidate feature lies closer to the vision-language query subspace, effectively capturing its structural geometric agreement with the joint constraints.

\subsubsection{Orientation Constraint}

The projection consistency score measures how much of the candidate feature can be explained by the query-induced anchor plane, but it does not specify whether the projected direction is supported by both query anchors. In a joint vision-language query, only the sector jointly indicated by the visual and textual anchors should be regarded as semantically valid. A candidate may obtain a large projection magnitude while its projected direction points away from this sector, causing the projection score to overestimate its compatibility with the joint query. To characterize this directional validity, we write the projection of $\mathbf{r}$ on the anchor plane as $\mathbf{p}=\alpha\mathbf{v}+\beta\mathbf{t}$, where $\alpha$ and $\beta$ denote the coordinates of the projection under the visual-textual anchor basis.

The coefficients $\alpha$ and $\beta$ are obtained from the orthogonal projection condition, which requires the residual $\mathbf{r}-\mathbf{p}$ to be orthogonal to both anchors. This leads to the following Gram system:
\begin{equation}
\begin{bmatrix}
1 & S_{vt}\\
S_{vt} & 1
\end{bmatrix}
\begin{bmatrix}
\alpha\\
\beta
\end{bmatrix}
=
\begin{bmatrix}
S_{vr}\\
S_{tr}
\end{bmatrix}.
\end{equation}
Solving this system gives
\begin{equation}
\begin{bmatrix}
\alpha\\
\beta
\end{bmatrix}
=
\frac{1}{1-S_{vt}^{2}}
\begin{bmatrix}
S_{vr}-S_{vt}S_{tr}\\
S_{tr}-S_{vt}S_{vr}
\end{bmatrix}.
\end{equation}
The non-negative cone $\{\alpha\mathbf{v}+\beta\mathbf{t}\mid \alpha\geq0,\beta\geq0\}$ defines the valid directed sector of the joint query. When both coefficients are non-negative, the projected candidate receives positive support from the visual and textual anchors. When either coefficient is negative, the projection falls outside this sector, indicating a directional conflict with at least one query cue.

For projections outside the valid directed sector, MAPS introduces an angular penalty according to the deviation from the nearest sector boundary. Since $\mathbf{v}$ and $\mathbf{t}$ are unit vectors and $\mathbf{p}$ is the orthogonal projection of $\mathbf{r}$ onto the anchor plane, we have $\mathbf{p}^{\top}\mathbf{v}=S_{vr}$ and $\mathbf{p}^{\top}\mathbf{t}=S_{tr}$. For $\|\mathbf{p}\|_2>0$, we denote the clipped nearest-boundary cosine as $c_{\mathrm{bd}}$, obtained by clipping $\max(S_{vr},S_{tr})/\|\mathbf{p}\|_2$ to $[-1,1]$. The deviation angle is then defined as
\begin{equation}
\theta_{\mathrm{dev}}
=
\begin{cases}
0,
& \alpha\geq0,\ \beta\geq0,\\
\arccos(c_{\mathrm{bd}}),
& \text{otherwise}.
\end{cases}
\end{equation}
If $\|\mathbf{p}\|_2=0$, the candidate has no valid projection on the anchor plane, and its MAPS score is set to zero.

The directional constraint is incorporated through a smooth angular decay, $W_{\mathrm{dir}}(\alpha,\beta)=\exp(-\theta_{\mathrm{dev}}^2)$. This weight equals one for projections inside the valid sector and decreases as the projected direction moves away from the nearest boundary. The final Multi-Anchor Projection Similarity is defined as
\begin{equation}
S_{\mathrm{MAPS}}(\mathbf{r};\mathbf{v},\mathbf{t})
=
S_{\mathrm{proj}}(\mathbf{r};\mathbf{v},\mathbf{t})
\cdot
W_{\mathrm{dir}}(\alpha,\beta).
\end{equation}
Geometrically, $S_{\mathrm{proj}}$ measures the target-to-subspace consistency between the candidate and the visual-textual anchor plane, while $W_{\mathrm{dir}}$ suppresses projections whose directions are not jointly supported by the two query anchors. MAPS therefore preserves the projection-based similarity formulation while enforcing directional validity under the joint vision-language constraint.
\subsection{MAPS-Based Contrastive Loss}

The MAPS score provides a geometry-aware similarity measure for joint vision-language retrieval. To make representation learning consistent with this retrieval criterion, we directly replace the pairwise cosine similarity in the contrastive objective \cite{contrastivelearning,infonce,CLIP,cpc} with the proposed MAPS score. In this way, the model is not optimized by independent image-reference or text-reference similarities, but by the consistency between each geo-referenced candidate and the anchor plane induced by the corresponding visual-textual query. For a mini-batch of $B$ training samples $\mathcal{B}=\{(\mathbf{v}_i,\mathbf{t}_i,\mathbf{r}_i)\}_{i=1}^{B}$, $\mathbf{v}_i$ and $\mathbf{t}_i$ denote the visual and textual query features of the $i$-th sample, while $\mathbf{r}_i$ denotes its matched geo-referenced candidate feature. For each joint query $(\mathbf{v}_i,\mathbf{t}_i)$, the positive reference is $\mathbf{r}_i$, while the remaining references in the mini-batch are treated as negatives. The query-to-reference MAPS contrastive loss is defined as
\begin{equation}
\mathcal{L}_{q\rightarrow r}
=
-\frac{1}{B}
\sum_{i=1}^{B}
\log
\frac{
\exp\left(
S_{\mathrm{MAPS}}(\mathbf{r}_i;\mathbf{v}_i,\mathbf{t}_i)/\tau
\right)
}{
\sum_{j=1}^{B}
\exp\left(
S_{\mathrm{MAPS}}(\mathbf{r}_j;\mathbf{v}_i,\mathbf{t}_i)/\tau
\right)
},
\end{equation}
where $\tau$ is the temperature parameter. This objective encourages the matched reference to obtain the highest MAPS response with respect to the corresponding joint query, while suppressing references that cannot be well explained by the same visual-textual anchor plane.

To further enforce one-to-one correspondence between joint queries and geo-referenced references, we apply the same MAPS-based objective in the reverse direction, following symmetric contrastive learning practices \cite{CLIP}:
\begin{equation}
\mathcal{L}_{r\rightarrow q}
=
-\frac{1}{B}
\sum_{j=1}^{B}
\log
\frac{
\exp\left(
S_{\mathrm{MAPS}}(\mathbf{r}_j;\mathbf{v}_j,\mathbf{t}_j)/\tau
\right)
}{
\sum_{i=1}^{B}
\exp\left(
S_{\mathrm{MAPS}}(\mathbf{r}_j;\mathbf{v}_i,\mathbf{t}_i)/\tau
\right)
}.
\end{equation}
Although MAPS is defined from a joint query to a candidate reference, this reverse loss does not change the geometric form of the similarity. It only requires each reference feature to be most consistent with its matched visual-textual query among all queries in the mini-batch. The final MAPS-based contrastive loss is calculated as
\begin{equation}
\mathcal{L}_{\mathrm{MAPS}}
=
\frac{1}{2}
\left(
\mathcal{L}_{q\rightarrow r}
+
\mathcal{L}_{r\rightarrow q}
\right).
\end{equation}
By directly optimizing MAPS in the contrastive objective \cite{contrastivelearning,infonce,CLIP,cpc}, the training process is aligned with the proposed retrieval geometry. Positive references are encouraged to lie close to the valid directed region of their corresponding visual-textual anchor plane, while negatives are penalized when they fail to satisfy the same joint-query constraint.

\section{Experiments}

\begin{table*}[!t]
\centering
\caption{Quantitative results of MAPS and existing methods on the CORE dataset. We report retrieval recall (R@1) and localization accuracy (L@150) on the global-level benchmark and four intercontinental subsets. Text-only and image-only denote single-modality query settings, while joint-query methods are evaluated under the UniMAG joint-query formulation. Bold indicates the best result, and underline indicates the second-best result.}
\label{tab:main_core}
\resizebox{\textwidth}{!}{%
\begin{tabular}{llcccccccccc}
\toprule
\multirow{3}{*}{\textbf{Method}} & \multirow{3}{*}{\textbf{Query}} & \multicolumn{2}{c}{\textbf{Global level}} & \multicolumn{8}{c}{\textbf{Intercontinental level}} \\
\cmidrule(lr){3-4} \cmidrule(lr){5-12}
 & & \multicolumn{2}{c}{All} & \multicolumn{2}{c}{Subset 1} & \multicolumn{2}{c}{Subset 2} & \multicolumn{2}{c}{Subset 3} & \multicolumn{2}{c}{Subset 4} \\
\cmidrule(lr){3-4} \cmidrule(lr){5-6} \cmidrule(lr){7-8} \cmidrule(lr){9-10} \cmidrule(lr){11-12}
 & & R@1 & L@150 & R@1 & L@150 & R@1 & L@150 & R@1 & L@150 & R@1 & L@150 \\
\midrule
CLIP-B/16 \cite{CLIP} & Text only & 40.11 & 44.16 & 42.18 & 44.74 & 43.77 & 46.96 & 37.32 & 42.37 & 37.39 & 42.73 \\
CLIP-L/14 \cite{CLIP} & Text only & 44.63 & 48.92 & 45.75 & 48.42 & 49.19 & 53.14 & 41.48 & 46.56 & 42.53 & 47.76 \\
BLIP \cite{blip} & Text only & 42.85 & 46.73 & 45.59 & 48.40 & 48.38 & 51.34 & 39.77 & 44.66 & 37.93 & 42.73 \\
CrossText2Loc \cite{CVGText} & Text only & 51.92 & 55.88 & 53.12 & 55.88 & 59.36 & 62.37 & 46.97 & 51.27 & 48.71 & 54.61 \\
PLANET \cite{CORE} & Text only & 55.84 & 59.66 & 57.90 & 60.72 & 64.74 & 67.58 & 49.81 & 54.13 & 51.35 & 56.65 \\
\midrule
CLIP-B/16 \cite{CLIP} & Image only & 59.93 & 62.92 & 62.26 & 64.82 & 66.81 & 69.18 & 53.37 & 57.95 & 57.93 & 60.16 \\
CLIP-L/14 \cite{CLIP} & Image only & 64.94 & 68.02 & 66.29 & 68.84 & 73.31 & 75.99 & 57.58 & 62.38 & 63.20 & 65.48 \\
BLIP \cite{blip} & Image only & 63.57 & 66.38 & 65.76 & 68.39 & 72.11 & 74.27 & 56.22 & 60.83 & 60.88 & 62.43 \\
\midrule
Pairwise alignment & Joint query & 70.16 & 73.05 & 74.21 & 76.66 & 77.18 & 79.55 & 62.05 & \underline{66.86} & 67.41 & 69.64 \\
VAST \cite{VAST} & Joint query & 69.42 & 72.27 & 73.42 & 75.88 & 75.92 & 78.71 & 61.57 & 65.90 & 66.22 & 68.83 \\
GRAM \cite{GRAM} & Joint query & 70.38 & 73.33 & 74.66 & 77.18 & 77.53 & 79.96 & 62.18 & 66.55 & 67.96 & 70.38 \\
PMRL \cite{PMRL} & Joint query & \underline{70.70} & \underline{73.65} & \underline{75.02} & \underline{77.46} & \underline{77.72} & \underline{80.12} & \underline{62.42} & 66.82 & \underline{68.48} & \underline{70.83} \\
MAPS & Joint query & \textbf{73.28} & \textbf{76.09} & \textbf{77.44} & \textbf{79.56} & \textbf{80.03} & \textbf{82.29} & \textbf{64.38} & \textbf{69.05} & \textbf{72.14} & \textbf{74.34} \\
\bottomrule
\end{tabular}%
}
\end{table*}

\begin{table*}[!t]
\centering
\caption{Quantitative results of MAPS and existing methods on the CVG-Text dataset. We report retrieval recall (R@1) and localization accuracy (L@50) on three city-level subsets, with both satellite-image and OSM reference galleries. Text-only and image-only denote single-modality query settings, while joint-query methods are evaluated under the UniMAG joint-query formulation. Bold indicates the best result, and underline indicates the second-best result. PLANET is not reported on OSM galleries since its image-based physical attribute mining is not applicable to symbolic map data.}
\label{tab:main_cvgtext}
\resizebox{\textwidth}{!}{%
\begin{tabular}{llcccccccccccc}
\toprule
\multirow{3}{*}{\textbf{Method}} & \multirow{3}{*}{\textbf{Query}} & \multicolumn{6}{c}{\textbf{Satellite image}} & \multicolumn{6}{c}{\textbf{OSM data}} \\
\cmidrule(lr){3-8} \cmidrule(lr){9-14}
 & & \multicolumn{2}{c}{New York} & \multicolumn{2}{c}{Brisbane} & \multicolumn{2}{c}{Tokyo} & \multicolumn{2}{c}{New York} & \multicolumn{2}{c}{Brisbane} & \multicolumn{2}{c}{Tokyo} \\
\cmidrule(lr){3-4} \cmidrule(lr){5-6} \cmidrule(lr){7-8} \cmidrule(lr){9-10} \cmidrule(lr){11-12} \cmidrule(lr){13-14}
 & & R@1 & L@50 & R@1 & L@50 & R@1 & L@50 & R@1 & L@50 & R@1 & L@50 & R@1 & L@50 \\
\midrule
CLIP-B/16 \cite{CLIP} & Text only & 26.67 & 28.17 & 29.92 & 32.58 & 24.00 & 27.25 & 27.42 & 29.42 & 30.83 & 32.67 & 17.75 & 19.92 \\
CLIP-L/14 \cite{CLIP} & Text only & 35.08 & 37.08 & 34.08 & 37.25 & 28.08 & 30.50 & 31.50 & 33.58 & 32.50 & 34.67 & 21.00 & 23.17 \\
BLIP \cite{blip} & Text only & 34.58 & 37.25 & 34.50 & 38.17 & 29.75 & 33.67 & 52.92 & 55.92 & 43.00 & 46.33 & 30.67 & 34.50 \\
CrossText2Loc \cite{CVGText} & Text only & 50.33 & 53.07 & 47.58 & 51.80 & 41.75 & 43.86 & 62.33 & 65.39 & 48.75 & 51.50 & 36.92 & 41.22 \\
PLANET \cite{CORE} & Text only & 52.83 & 55.58 & 49.83 & 53.67 & 43.42 & 46.42 & -- & -- & -- & -- & -- & -- \\
\midrule
CLIP-B/16 \cite{CLIP} & Image only & 67.58 & 69.42 & 68.92 & 71.83 & 62.33 & 65.25 & 20.83 & 23.17 & 22.75 & 25.08 & 12.92 & 15.33 \\
CLIP-L/14 \cite{CLIP} & Image only & 72.75 & 74.75 & 74.25 & 77.50 & 67.92 & 70.67 & 24.58 & 27.00 & 25.42 & 27.92 & 15.83 & 18.33 \\
BLIP \cite{blip} & Image only & 71.33 & 73.50 & 73.08 & 76.08 & 66.25 & 69.08 & 28.67 & 29.75 & 27.92 & 29.83 & 20.83 & 24.08 \\
\midrule
Pairwise alignment & Joint query & \underline{79.33} & \underline{80.67} & \underline{82.17} & \underline{85.08} & \underline{71.67} & \underline{75.33} & \underline{64.42} & \underline{67.25} & \underline{50.33} & \underline{53.08} & \underline{38.92} & \underline{43.00} \\
VAST \cite{VAST} & Joint query & 76.84 & 78.22 & 79.64 & 82.70 & 68.92 & 72.31 & 61.38 & 64.52 & 47.21 & 50.46 & 35.94 & 40.18 \\
GRAM \cite{GRAM} & Joint query & 78.10 & 79.46 & 80.58 & 83.64 & 70.24 & 73.68 & 62.81 & 65.73 & 48.64 & 51.57 & 37.26 & 41.03 \\
PMRL \cite{PMRL} & Joint query & 77.62 & 79.03 & 81.05 & 84.11 & 69.81 & 74.02 & 62.46 & 66.14 & 49.08 & 52.06 & 36.88 & 41.62 \\
MAPS & Joint query & \textbf{84.00} & \textbf{85.42} & \textbf{86.17} & \textbf{88.42} & \textbf{75.92} & \textbf{79.17} & \textbf{67.83} & \textbf{70.17} & \textbf{52.58} & \textbf{55.33} & \textbf{40.75} & \textbf{45.08} \\
\bottomrule
\end{tabular}%
}
\end{table*}

\subsection{Experimental Setup}

\subsubsection{Datasets and Protocols}

We evaluate UniMAG on two  geo-localization datasets that contain both visual observations and natural-language descriptions, namely CORE \cite{CORE} and CVG-Text \cite{CVGText}. For both datasets, a query is formed by a ground-level image and its associated textual description, and the goal is to retrieve the geo-referenced candidate corresponding to the same physical location from the gallery. 

\textit{CORE.} CORE is a million-scale benchmark for global cross-modal geo-localization, containing 1,034,786 multi-view images with fine-grained textual annotations from 225 representative geographic regions across six continents. It covers diverse urban, suburban, natural, and built environments, and integrates street-level observations with high-resolution satellite imagery for multi-scale localization evaluation. This broad geographical, scale, and scene diversity makes CORE a challenging benchmark for testing whether MAPS can exploit complementary visual-textual cues under large-scale spatial variation. 

\textit{CVG-Text.} CVG-Text is a city-level CMGL dataset introduced by Ye et al. \cite{CVGText}, containing 30,000 scene points from three major cities: New York, Brisbane, and Tokyo. It uses GPT-4o to generate descriptive texts for ground-level observations, and provides both satellite imagery and OpenStreetMap (OSM) data as geo-referenced candidate galleries. Compared with the global-scale setting of CORE, CVG-Text has a narrower geographic span and primarily focuses on modern metropolitan environments, making it a complementary benchmark for evaluating MAPS under city-scale vision-language retrieval with different reference modalities. We follow the official data organization and construct joint queries from the paired ground-view images and text descriptions.

\subsubsection{Baselines}

The baselines are designed to separate the benefit of additional query modalities from the benefit of the proposed multi-anchor geometry. First, we include single-modality retrieval baselines, where only the image query or only the text query is used. These baselines include CLIP-B/16 \cite{CLIP}, CLIP-L/14 \cite{CLIP}, and BLIP \cite{blip} variants when applicable, and they indicate how much localization evidence can be obtained from each modality alone. We also include specific CMGL methods based on CLIP-L/14, including CrossText2Loc \cite{CVGText} and PLANET \cite{CORE}. Since these methods are designed for text-guided geo-localization rather than joint vision-language queries, they are reported as text-query baselines instead of direct VLGL competitors.

To examine the effect of the alignment strategy itself, we keep the backbone setting balanced in the controlled comparison. Specifically, pairwise alignment and MAPS both use the OpenAI pre-trained CLIP-ViT-L/14@336px model \cite{CLIP,vit} as the image and text backbone. The image query encoder and reference encoder share the same visual backbone, while the text encoder uses the corresponding CLIP text branch. We further compare with representative multimodal alignment methods, including VAST \cite{VAST}, GRAM \cite{GRAM}, and PMRL \cite{PMRL}, by adapting them to the same joint-query retrieval setting. For a fair comparison, experiments on similarity measures and training objectives use the same backbone, candidate gallery, query split, and fine-tuning data.

\subsubsection{Evaluation Metrics}

We report retrieval accuracy and localization accuracy, following common geo-localization evaluation protocols \cite{CVUSA,CVACT,vigor,Sample4geo,CVGText,CORE}. For retrieval, we use top-1 recall (R@1), which measures the percentage of queries for which the ground-truth reference is ranked first. For localization, we report the success rate within a dataset-specific distance threshold, denoted as L@150 on CORE and L@50 on CVG-Text. We use the larger threshold for CORE because it covers broader geographic regions, while CVG-Text is evaluated at a finer city scale. Retrieval accuracy and localization accuracy together measure whether a method can rank the correct reference first and produce a geographically valid top prediction.

\subsubsection{Implementation Details}

All models are fine-tuned on the corresponding dataset before evaluation. Before feature extraction, input images are resized to $336\times336$, and text descriptions are tokenized with a context length of $N=300$. All output features are $L_2$-normalized before similarity computation. We train each model for 40 epochs with a batch size of 64. Optimization is performed with Adam \cite{adam}, using an initial learning rate of $1\times10^{-5}$ and cosine learning rate decay. All experiments are conducted on four NVIDIA RTX 4090 GPUs. For MAPS, we add a small numerical regularizer to the Gram determinant of the anchor plane to avoid numerical instability when the anchor vectors become nearly collinear.

\begin{table*}[t]
\centering
\caption{Ablation results of different MAPS components on CORE. This table compares cosine fusion, MAPS with only the projection score, and the full MAPS with the direction-aware constraint, showing the contributions of projection consistency and directional validity.}
\label{tab:component_ablation}
\setlength{\tabcolsep}{4pt}
\resizebox{0.78\textwidth}{!}{%
\begin{tabular}{lcccccccc}
\toprule
\multirow{2}{*}{\textbf{Method}} & \multicolumn{2}{c}{\textbf{Subset 1}} & \multicolumn{2}{c}{\textbf{Subset 2}} & \multicolumn{2}{c}{\textbf{Subset 3}} & \multicolumn{2}{c}{\textbf{Subset 4}} \\
\cmidrule(lr){2-3} \cmidrule(lr){4-5} \cmidrule(lr){6-7} \cmidrule(lr){8-9}
 & R@1 & L@150 & R@1 & L@150 & R@1 & L@150 & R@1 & L@150 \\
\midrule
Cosine fusion & 74.21 & 76.66 & 77.18 & 79.55 & 62.05 & 66.86 & 67.41 & 69.64 \\
Projection-only MAPS ($S_{\mathrm{proj}}$) & 77.21 & 79.08 & 79.72 & 81.85 & 64.11 & 68.66 & 71.68 & 74.12 \\
Direction-aware MAPS & \textbf{77.44} & \textbf{79.56} & \textbf{80.03} & \textbf{82.29} & \textbf{64.38} & \textbf{69.05} & \textbf{72.14} & \textbf{74.34} \\
\bottomrule
\end{tabular}%
}
\end{table*}

\begin{table*}[t]
\centering
\caption{Analysis of the role of MAPS in training and retrieval. This table compares conventional pairwise alignment, MAPS used only during test-time retrieval, and MAPS used in both training and test-time retrieval, showing the effect of MAPS as a retrieval score and as a training objective.}
\label{tab:maps_usage_ablation}
\setlength{\tabcolsep}{4pt}
\resizebox{0.78\textwidth}{!}{%
\begin{tabular}{lcccccccc}
\toprule
\multirow{2}{*}{\textbf{Method}} & \multicolumn{2}{c}{\textbf{Subset 1}} & \multicolumn{2}{c}{\textbf{Subset 2}} & \multicolumn{2}{c}{\textbf{Subset 3}} & \multicolumn{2}{c}{\textbf{Subset 4}} \\
\cmidrule(lr){2-3} \cmidrule(lr){4-5} \cmidrule(lr){6-7} \cmidrule(lr){8-9}
 & R@1 & L@150 & R@1 & L@150 & R@1 & L@150 & R@1 & L@150 \\
\midrule
Pairwise alignment & 74.21 & 76.66 & 77.18 & 79.55 & 62.05 & 66.86 & 67.41 & 69.64 \\
MAPS for retrieval only & 76.18 & 78.47 & 78.36 & 80.74 & 63.21 & 67.74 & 69.19 & 71.44 \\
\textbf{MAPS for optimization and retrieval} & \textbf{77.44} & \textbf{79.56} & \textbf{80.03} & \textbf{82.29} & \textbf{64.38} & \textbf{69.05} & \textbf{72.14} & \textbf{74.34} \\
\bottomrule
\end{tabular}%
}
\end{table*}

\subsection{Comparison with State of the Art}

\subsubsection{Results on CORE}

To evaluate the effectiveness of MAPS on large-scale VLGL, Table~\ref{tab:main_core} reports the results on the global-level benchmark and four intercontinental-level subsets of CORE. These comparisons cover both single-modality baselines and joint-query methods, allowing us to distinguish the benefit of using additional query modalities from the benefit of explicitly modeling the multi-anchor geometry.

The results show that image-only retrieval generally performs better than text-only retrieval on CORE, suggesting that ground-view images provide stronger location-discriminative evidence in large-scale geo-localization. Meanwhile, joint-query methods further improve retrieval performance by incorporating textual semantic constraints, demonstrating the complementarity between visual and textual cues. However, conventional multimodal alignment strategies still rely on pairwise similarity aggregation or global representation alignment, and therefore do not fully exploit the geometric structure jointly defined by the visual and textual queries.

Compared with these baselines, UniMAG + MAPS achieves the best performance across all metrics and geographic partitions. On the global-level benchmark, it improves over pairwise alignment by 3.12\% in R@1 and 3.04\% in L@150. It also surpasses the strongest geometry-aware multimodal alignment baseline, PMRL, by 2.58\% and 2.44\%, respectively. Across the four intercontinental-level subsets, UniMAG + MAPS consistently ranks first, further confirming its robustness under diverse geographic environments. These results show that MAPS provides a more faithful retrieval principle for VLGL. By scoring candidates according to their consistency with the visual-textual query subspace, MAPS preserves cross-modal complementarity and yields more discriminative localization than either pairwise fusion or global modality alignment.

\subsubsection{Results on CVG-Text}

To further evaluate the generalization ability of UniMAG under different localization scales and reference modalities, we conduct experiments on CVG-Text. Table~\ref{tab:main_cvgtext} reports the results on three city-level subsets, including New York, Brisbane, and Tokyo, with both satellite-image and OSM reference galleries. CVG-Text also contains two different reference modalities, allowing us to examine whether MAPS remains effective when the retrieval target changes from visual satellite imagery to symbolic map data.

The results show different modality preferences across the two reference galleries. For the satellite-image gallery, image-only retrieval performs much better than text-only retrieval, indicating that visual appearance is the dominant cue when the reference candidates are satellite images. Joint-query methods further improve the results by incorporating textual semantic constraints, showing that language can help disambiguate visually similar locations. For the OSM gallery, the trend is different. Since OSM data mainly encodes map symbols, road topology, and POI-related semantics rather than raw visual appearance \cite{CVGText}, text-only retrieval becomes stronger than image-only retrieval. This contrast provides effective support for evaluating MAPS in a text-dominant retrieval environment.

Compared with pairwise alignment, existing multimodal alignment methods remain slightly lower on CVG-Text. This is mainly due to the fitting limitation of global alignment objectives on a small-scale city-level dataset. Methods such as GRAM and PMRL rely on high-order global alignment \cite{GRAM,PMRL}, which tends to push multimodal features toward strong collinearity. Such objectives usually require sufficient training samples and geographic diversity to learn a stable global geometric structure. In contrast, pairwise alignment directly optimizes the observed query-reference relations and is therefore easier to fit, making it a strong baseline on CVG-Text. MAPS introduces a more task-oriented geometric constraint. Instead of forcing modality features to collapse into a single direction, it encourages the target feature to remain consistent with the anchor plane jointly defined by the visual and textual queries. As a result, across the six gallery settings, UniMAG + MAPS improves over pairwise alignment by 1.83\%--4.67\% in R@1 and 2.08\%--4.75\% in L@50, achieving the best performance among all compared methods.

\subsection{Ablation Studies}

\subsubsection{Analysis of MAPS Scoring Components}

We first analyze the scoring components of MAPS to examine how each geometric term contributes to retrieval performance. MAPS consists of two key components: the projection response $S_{\mathrm{proj}}$, which measures the consistency between a candidate feature and the visual-textual anchor plane, and the direction-aware constraint, which suppresses projections falling outside the valid sector jointly supported by the two query anchors. To isolate their effects, we compare three ranking strategies on the four intercontinental subsets of CORE: conventional cosine fusion, projection-only MAPS, and direction-aware MAPS.

As shown in Table~\ref{tab:component_ablation}, replacing cosine fusion with projection-only MAPS consistently improves both retrieval and localization performance across all CORE subsets. Compared with cosine fusion, projection-only MAPS improves R@1 by 2.06\%--4.27\% and L@150 by 1.80\%--4.48\%. This result indicates that measuring a candidate by its projection on the visual-textual anchor plane is more effective than aggregating two isolated pairwise cosine similarities. It also supports our formulation that a joint vision-language query should be modeled as a query-induced subspace rather than as two independent retrieval anchors.

The direction-aware constraint further improves the projection-only variant on all subsets. Although the gain is relatively moderate, it is consistent, improving R@1 by 0.23\%--0.46\% and L@150 by 0.22\%--0.48\%. This observation is consistent with the role of the direction-aware term. The projection response captures the main target-to-subspace consistency, while the directional constraint corrects candidates that have large projection magnitudes but fall outside the valid sector jointly indicated by the visual and textual anchors. Therefore, the full direction-aware MAPS achieves the best performance among all scoring variants, demonstrating that projection consistency and directional validity are complementary components in the proposed similarity metric.

\subsubsection{Effect of MAPS in Retrieval and Optimization}

We then analyze the role of MAPS in different stages of the learning and retrieval pipeline. Specifically, MAPS can be used either only as an inference-time ranking score or as both the training objective and the retrieval metric. This distinction is important because using MAPS only during inference tests the discriminative ability of the proposed similarity on representations learned by pairwise contrastive training, while using MAPS in both optimization and retrieval tests whether training-inference consistency further improves the learned embedding space.

Table~\ref{tab:maps_usage_ablation} reports the comparison among conventional pairwise alignment, MAPS used only for retrieval, and MAPS used for both optimization and retrieval. The retrieval-only setting already improves over pairwise alignment, increasing R@1 by 1.16\%--1.97\% and L@150 by 0.88\%--1.81\% across the CORE subsets. This confirms that MAPS itself provides a stronger ranking criterion than pairwise cosine fusion, even when the representation is learned with a conventional pairwise contrastive objective.

Using MAPS in both optimization and retrieval further improves the retrieval-only setting by 1.17\%--2.95\% in R@1 and 1.09\%--2.90\% in L@150. This result shows that MAPS is not only useful as a post-hoc retrieval metric, but also beneficial as a training objective. When the model is optimized with the same target-to-subspace relation used for final ranking, the learned representation becomes more consistent with the multi-anchor retrieval geometry. As a result, MAPS for optimization and retrieval achieves the best performance across all subsets, improving over pairwise alignment by 2.33\%--4.73\% in R@1 and 2.19\%--4.70\% in L@150.

\subsection{Efficiency Analysis}

\begin{figure}[!t]
\centering
\includegraphics[width=\columnwidth]{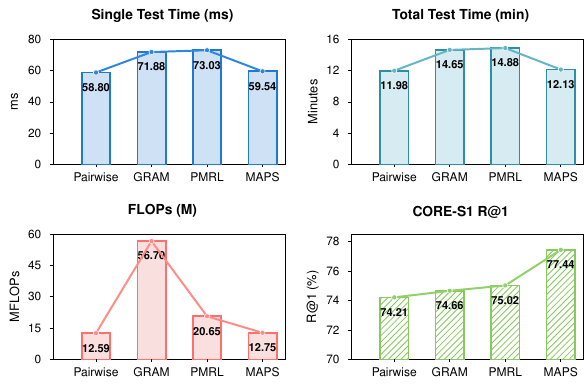}
\caption{Comparison of single-query testing time, total testing time, scoring FLOPs, and retrieval performance for four methods on CORE-S1.}
\label{fig:efficiency_comparison}
\end{figure}

\begin{figure}[!t]
\centering
\includegraphics[width=\columnwidth]{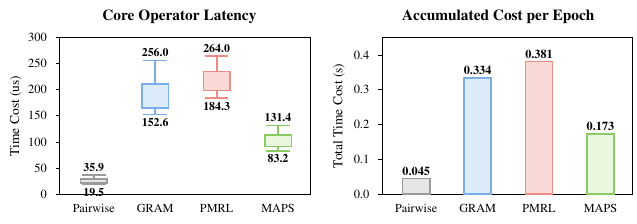}
\caption{Core operator cost comparison among Pairwise alignment, GRAM, PMRL, and MAPS. The left panel reports the batch-level latency of each scoring operator measured with a mini-batch. The right panel reports the accumulated operator cost for one epoch.}
\label{fig:operator_cost}
\end{figure}

To evaluate its practical viability, we compare MAPS with pairwise alignment, GRAM, and PMRL on CORE Subset 1, where all methods use the same backbone and evaluation setting. The results are shown in Fig.~\ref{fig:efficiency_comparison} and Fig.~\ref{fig:operator_cost}.

Compared with pairwise alignment, MAPS introduces only very limited additional computational overhead, with the scoring FLOPs and total testing time increasing by only 1.27\% and 1.25\%, respectively. This is because MAPS only requires a small number of scalar operations to obtain the projection length and directional constraint. Compared with existing multimodal alignment methods, MAPS also shows higher computational efficiency. Its total testing time is reduced by 17.18\% and 18.48\% compared with GRAM and PMRL, respectively. Fig.~\ref{fig:operator_cost} further isolates the computational cost of the similarity scoring operator, excluding shared procedures such as feature encoding, data loading, and distributed communication. When only the similarity scoring stage during training is considered, the average mini-batch latency of MAPS is only 51.89\% of GRAM and 45.56\% of PMRL. This advantage comes from the task-specific geometric modeling of MAPS. In the two-anchor VLGL setting, MAPS can directly express the relation between the target feature and the visual-textual anchor plane through a closed-form projection, thereby avoiding more expensive high-order matrix operations used in general multimodal alignment methods \cite{GRAM,PMRL}. Together with the retrieval results on CORE-S1, these comparisons show that MAPS can enlarge the performance margin while keeping the computational cost close to pairwise alignment. This indicates that its gain mainly comes from a similarity definition better suited to VLGL, rather than from heavier computational resources.

\subsection{Geometric Interpretability Analysis}

\begin{figure*}[!t]
\centering
\includegraphics[width=0.98\textwidth]{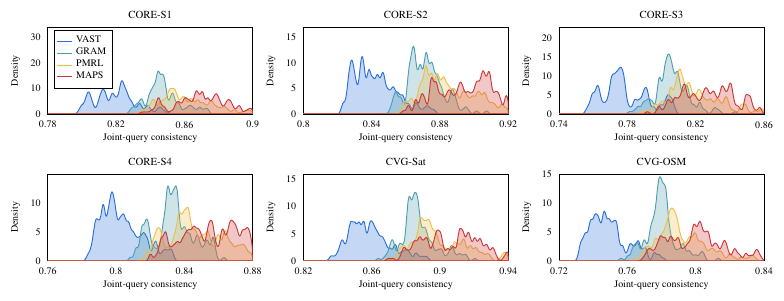}
\caption{Comparison of joint-query consistency distributions for different multimodal alignment methods, indicating the stronger alignment of MAPS with the visual-textual query subspace.}
\label{fig:joint_query_consistency}
\end{figure*}

\textbf{Joint-query consistency distribution.}
To examine the local geometry of the representations learned by MAPS, Fig.~\ref{fig:joint_query_consistency} compares the joint-query consistency distributions of different methods on the four intercontinental subsets of CORE and the two reference galleries of CVG-Text. This analysis is conducted offline and is not involved in MAPS training or inference. Specifically, we construct a local representation matrix $\mathbf{Z}=[\mathbf{v},\mathbf{t},\mathbf{r}]$ from the visual query, textual query, and candidate reference feature, and compute the ratio between the energy of the top two singular values and the total singular-value energy \cite{PMRL,GRAM}. This statistic characterizes the extent to which the triplet can be explained by the two-dimensional subspace spanned by the visual and textual anchors. Across all evaluation settings, MAPS produces distributions that are consistently shifted to the right, suggesting stronger local consistency between candidate references and the joint-query subspace. In contrast, VAST, GRAM, and PMRL exhibit relatively left-shifted distributions, suggesting that global multimodal alignment is not always sufficient to induce strong query-candidate subspace consistency. This result provides complementary geometric evidence that MAPS better fits the joint-query constraint in VLGL.

\begin{figure*}[!t]
\centering
\includegraphics[width=0.98\textwidth]{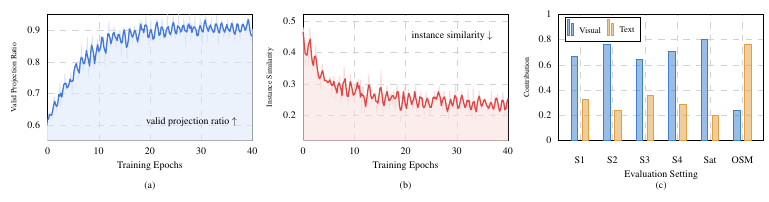}
\caption{Training dynamics and modality contribution analysis of MAPS. Subfigure (a) shows the increasing valid projection ratio during training. Subfigure (b) shows the decreasing instance similarity. Subfigure (c) decomposes the MAPS score into visual-anchor and textual-anchor contributions across different evaluation settings.}
\label{fig:maps_training_contribution}
\end{figure*}

\textbf{Valid projection ratio.}
Fig.~\ref{fig:maps_training_contribution}(a) presents the training dynamics of the valid projection ratio, which measures whether the projection of a candidate reference falls within the effective region jointly supported by the visual and textual anchors. The ratio gradually increases and then becomes stable as training proceeds, indicating that the MAPS-based objective progressively improves the consistency between positive references and the valid joint-query region. This phenomenon shows that the role of MAPS is not limited to an inference-time ranking criterion. Its training objective also guides the embedding space toward a geometry that is more suitable for joint vision-language retrieval.

\textbf{Instance similarity.}
For VLGL, different geographic locations often share similar road structures, architectural appearances, or semantic descriptions. Therefore, the model needs to preserve fine-grained location separability while enforcing cross-modal consistency. Fig.~\ref{fig:maps_training_contribution}(b) analyzes the training dynamics of instance similarity. The overall decrease in similarity between different instances suggests that MAPS improves target-to-subspace consistency without compressing all samples into the same shared direction, thereby preserving the necessary instance-level discriminability.

\textbf{Modality contribution interpretation.}
Fig.~\ref{fig:maps_training_contribution}(c) decomposes the contributions of the visual and textual anchors in the MAPS score. On CORE-S1 to CORE-S4 and CVG-Sat, the visual anchor contributes more, which is consistent with the satellite-image reference setting where visual appearance matching is dominant. In contrast, the textual anchor becomes dominant on CVG-OSM, matching the nature of OSM references that mainly encode road topology, map symbols, and POI semantics rather than raw visual appearance. This result indicates that MAPS supports modality-adaptive joint-query matching, with the relative contributions of visual and textual anchors varying according to the reference modality and retrieval environment.

\subsection{Robustness Analysis}

In practical VLGL scenarios, image queries may be affected by occlusion, blur, or viewpoint changes \cite{lowry2015visual,CVACT,vigor}, while textual queries may contain incomplete descriptions or semantic noise. Such perturbations can disturb the stability of query features and affect candidate-reference matching. To evaluate the robustness of MAPS under noisy query conditions, we add Gaussian noise with a scale of 0.4 to the normalized query image feature and query text feature, respectively \cite{noise}. Table~\ref{tab:noise_robustness} reports the results on CORE-S1. All methods suffer performance degradation after noise is introduced, while MAPS still achieves the highest R@1 and L@150 under both perturbation settings. Compared with pairwise alignment, GRAM, and PMRL, MAPS exhibits a smaller performance drop, indicating that its multi-anchor projection mechanism can exploit the uncorrupted query modality as a complementary constraint, thereby improving robustness under noisy query conditions in VLGL.

\begin{table*}[!t]
\centering
\caption{Robustness comparison with Gaussian noise added to image and text query features on CORE-S1. Values in parentheses denote performance drops from the clean CORE-S1 results.}
\label{tab:noise_robustness}
\small
\setlength{\tabcolsep}{3.5pt}
\resizebox{0.75\textwidth}{!}{%
\begin{tabular}{lcccccccc}
\toprule
\multirow{2}{*}{\textbf{Noise type}} &
\multicolumn{2}{c}{\textbf{Pairwise}} &
\multicolumn{2}{c}{\textbf{GRAM}} &
\multicolumn{2}{c}{\textbf{PMRL}} &
\multicolumn{2}{c}{\textbf{MAPS}} \\
\cmidrule(lr){2-3} \cmidrule(lr){4-5} \cmidrule(lr){6-7} \cmidrule(lr){8-9}
& R@1 & L@150 & R@1 & L@150 & R@1 & L@150 & R@1 & L@150 \\
\midrule
Image-query noise & 59.16{\scriptsize\,($\downarrow$15.05)} & 61.74{\scriptsize\,($\downarrow$14.92)} & 62.08{\scriptsize\,($\downarrow$12.58)} & 64.31{\scriptsize\,($\downarrow$12.87)} & 63.73{\scriptsize\,($\downarrow$11.29)} & 66.02{\scriptsize\,($\downarrow$11.44)} & \textbf{68.41}{\scriptsize\,($\downarrow$9.03)} & \textbf{70.18}{\scriptsize\,($\downarrow$9.38)} \\
Text-query noise & 63.01{\scriptsize\,($\downarrow$11.20)} & 65.64{\scriptsize\,($\downarrow$11.02)} & 65.92{\scriptsize\,($\downarrow$8.74)} & 67.38{\scriptsize\,($\downarrow$9.80)} & 67.11{\scriptsize\,($\downarrow$7.91)} & 69.47{\scriptsize\,($\downarrow$7.99)} & \textbf{72.36}{\scriptsize\,($\downarrow$5.08)} & \textbf{74.09}{\scriptsize\,($\downarrow$5.47)} \\
\bottomrule
\end{tabular}
}
\end{table*}

\subsection{Cross-Domain Generalization}

To evaluate the generalization ability of MAPS in unseen geographic regions \cite{vigor,CVCITIES,CORE}, we conduct cross-subset generalization experiments on CORE. Specifically, we use Subset 4 as the target testing domain and train the model separately on Subset 1, Subset 2, and Subset 3.

\begin{table}[!t]
\centering
\caption{Quantitative comparison of out-of-distribution generalization on the CORE dataset.}
\label{tab:generalization}
\resizebox{\columnwidth}{!}{%
\begin{tabular}{lcccccc}
\toprule
\multirow{2}{*}{\textbf{Method}} & \multicolumn{2}{c}{\textbf{Subset 1 $\rightarrow$ 4}} & \multicolumn{2}{c}{\textbf{Subset 2 $\rightarrow$ 4}} & \multicolumn{2}{c}{\textbf{Subset 3 $\rightarrow$ 4}} \\
\cmidrule(lr){2-3} \cmidrule(lr){4-5} \cmidrule(lr){6-7}
 & R@1 & L@150 & R@1 & L@150 & R@1 & L@150 \\
\midrule
Pairwise alignment & 49.78 & 52.03 & 48.91 & 51.33 & 55.98 & 58.28 \\
VAST \cite{VAST} &
47.25 & 49.08 &
46.83 & 47.75 &
53.08 & 56.17 \\
GRAM \cite{GRAM} & 48.74 & 51.31 & 46.56 & 49.48 & 54.09 & 56.02 \\
PMRL \cite{PMRL} & 50.41 & 53.06 & 49.88 & 51.94 & 56.52 & 59.41 \\
\midrule
MAPS & \textbf{54.36} & \textbf{57.22} & \textbf{53.91} & \textbf{55.74} & \textbf{60.62} & \textbf{63.58} \\
\bottomrule
\end{tabular}
}
\end{table}

\begin{figure*}[!htbp]
\centering
\includegraphics[width=6.4 in]{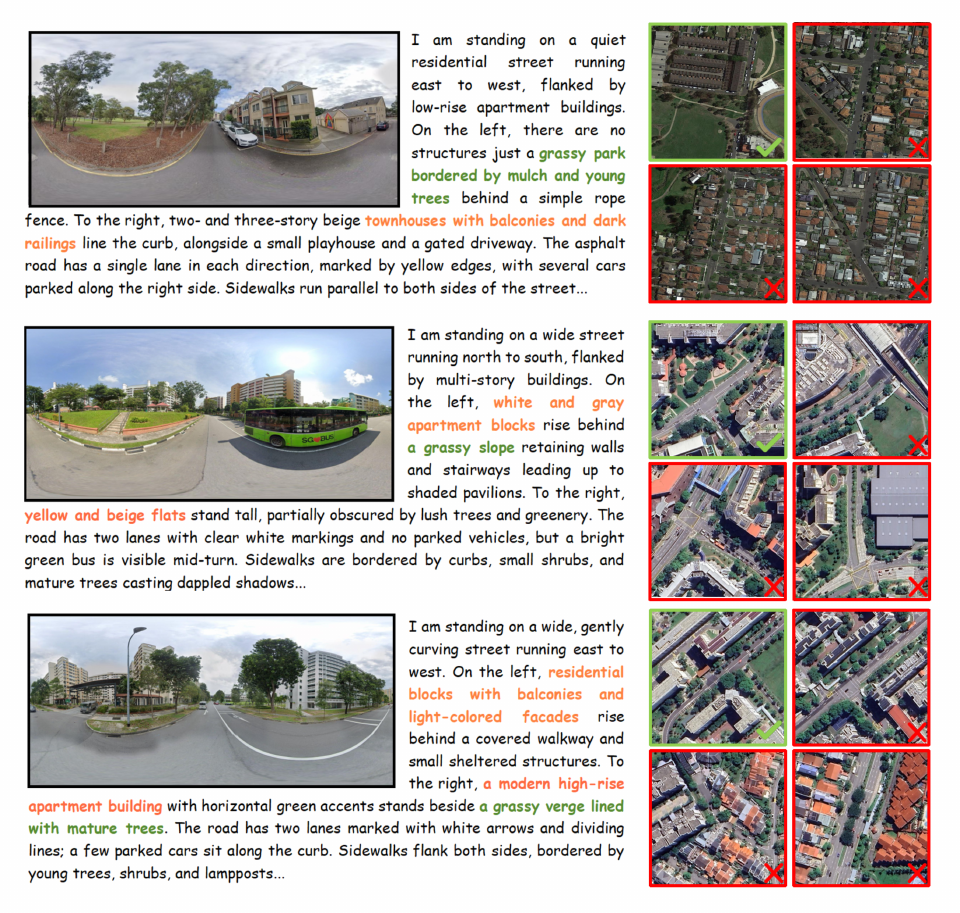}
\caption{The retrieval results of MAPS on the CORE dataset. The left side displays the original street-view image together with its corresponding textual description. The right side illustrates the top-4 retrieval results, where green highlights indicate correct matches and red highlights indicate incorrect results.}
\label{VISUAL}
\end{figure*}

\begin{figure*}[!htbp]
\centering
\includegraphics[width=6.4 in]{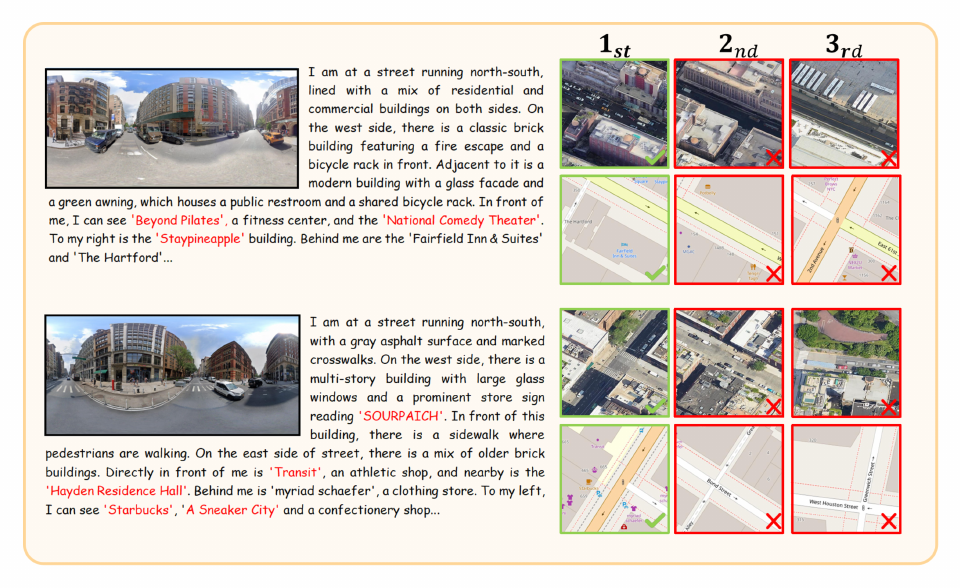}
\caption{The retrieval results of MAPS on the CVG-Text dataset. The left side displays the original street-view image together with its corresponding textual description. The right side illustrates the top-3 retrieval results, where green highlights indicate correct matches and red highlights indicate incorrect results.}
\label{CVG_VISUAL}
\end{figure*}

Table~\ref{tab:generalization} reports the cross-domain generalization results. All methods exhibit clear performance degradation under cross-subset evaluation, indicating that geographic distribution shifts pose a substantial challenge to VLGL. Nevertheless, MAPS achieves the best performance in all three transfer settings, demonstrating more stable retrieval capability under domain shift. Compared with pairwise alignment and existing multimodal alignment methods, the projection mechanism of MAPS places stronger emphasis on the consistency between candidate references and the visual-textual query subspace. Since this subspace is jointly defined by two complementary query anchors, a candidate can obtain a high response only when it satisfies both visual structural constraints and textual semantic constraints. As a result, MAPS suppresses domain-specific matches supported by only a single modality and maintains more stable ranking results in cross-domain scenarios. These results suggest that the multi-anchor geometric retrieval criterion provides a more transferable ranking principle for VLGL.

\subsection{Qualitative Results and Visualization}

\begin{figure}[!htbp]
\centering
\includegraphics[width=\columnwidth]{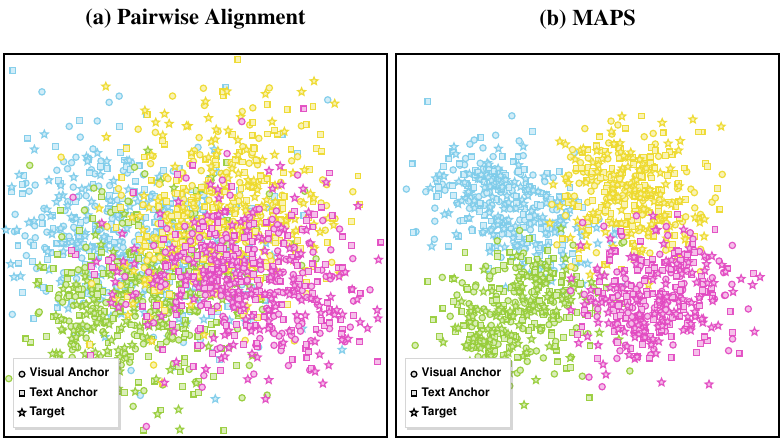}
\caption{t-SNE visualization \cite{t-SNE} of the learned CORE embedding space under pairwise alignment and MAPS. Colors denote different geographic subsets, while circles, squares, and stars indicate visual anchors, textual anchors, and target references, respectively. MAPS yields more coherent subset-wise structures while preserving the joint-query retrieval geometry.}
\label{fig:tsne_visualization}
\end{figure}

Fig.~\ref{VISUAL} and Fig.~\ref{CVG_VISUAL} present representative VLGL retrieval results of MAPS on CORE and CVG-Text. Fig.~\ref{VISUAL} shows that MAPS can jointly exploit road orientation, building layout, vegetation distribution, and fine-grained textual cues such as residential buildings, trees, grassy areas, and road structures to identify the correct satellite reference among visually similar candidates. Fig.~\ref{CVG_VISUAL} further evaluates MAPS under both satellite-image and OSM reference galleries. In the OSM setting, POI names, street relations, and landmark semantics in the text provide crucial localization constraints, enabling MAPS to retrieve the correct location from symbolic map representations. To further examine the learned representation structure, Fig.~\ref{fig:tsne_visualization} visualizes the CORE embedding space under pairwise alignment and MAPS. Compared with the relatively mixed distribution produced by pairwise alignment, MAPS forms clearer subset-wise structures and places target references within the region jointly constrained by visual and textual anchors. These qualitative results indicate that MAPS can adaptively integrate visual structural cues and textual semantic cues, yielding a more coherent embedding space and more reliable VLGL retrieval across different reference modalities.

\section{Limitations}

Although MAPS achieves strong retrieval performance and computational efficiency in VLGL, several directions remain open for further investigation. First, this work mainly focuses on the two-anchor setting formed by an image query and a text query, where the visual and textual anchors jointly define the query subspace. For scenarios involving more than two query modalities or multiple query instances, how to construct a stable and interpretable high-dimensional multi-anchor subspace requires further study. Second, the effectiveness of MAPS relies on the assumption that different query modalities provide complementary and reasonably reliable localization cues. When one modality is missing or semantically inconsistent with another modality, the geometric structure defined by the query anchors may become unstable, which can affect candidate-reference matching. In addition, maintaining retrieval robustness when the available modality combinations differ between training and testing remains an open problem. Future work will explore more general multi-anchor extensions, modality reliability modeling, and adaptive retrieval mechanisms for modality-missing and modality-shifted VLGL settings.

\section{Conclusion}

This paper studies VLGL with joint image-text queries and introduces UniMAG as a unified framework for this retrieval setting. Built upon this framework, we propose MAPS, a multi-anchor projection similarity that treats visual and textual queries as coordinated anchors and evaluates each geo-referenced candidate by its consistency with the visual-textual query subspace. Unlike isolated pairwise similarity, MAPS combines target-to-subspace projection with an orientation-aware constraint, capturing both the complementary support and directional validity of joint query cues. We further incorporate MAPS into the contrastive learning objective, aligning representation learning with the retrieval geometry used at inference time. Experiments on two datasets show that MAPS consistently improves VLGL performance across different localization scales, geographic regions, and reference modalities, while adding only marginal computational overhead over pairwise alignment. These results demonstrate that multi-anchor geometric matching provides a principled and practical retrieval criterion for joint-query geo-localization, and offers a useful perspective for modeling more flexible multimodal localization systems.

\bibliographystyle{IEEEtran} 
\bibliography{reference}

\end{document}